\documentclass[10pt,twocolumn,letterpaper]{article}

\usepackage[pagenumbers]{cvpr} 

\usepackage{microtype}
\usepackage{multirow}
\usepackage{array}
\renewcommand{\paragraph}[1]{\vspace{.5em}\noindent\textbf{#1.}}

\usepackage{tikz}
\usepackage{pgfplots}
\pgfplotsset{compat=1.18}

\definecolor{cvprblue}{rgb}{0.21,0.49,0.74}
\usepackage[pagebackref,breaklinks,colorlinks,allcolors=cvprblue]{hyperref}

\def\paperID{*****} 
\def\confName{CVPR}
\def\confYear{2026}

\title{Grounding Synthetic Data Generation With Vision and Language Models}
\author{Ümit Mert Çağlar, Alptekin Temizel\\
Graduate School of Informatics\\
METU, Turkey\\
{\tt\small mecaglar@metu.edu.tr, atemizel@metu.edu.tr}
}

\begin{document}
\maketitle
\begin{abstract}
Deep learning models benefit from increasing data diversity and volume, motivating synthetic data augmentation to improve existing datasets. However, existing evaluation metrics for synthetic data typically calculate latent feature similarity, which is difficult to interpret and does not always correlate with the contribution to downstream tasks.

We propose a vision-language grounded framework for interpretable synthetic data augmentation and evaluation in remote sensing. Our approach combines generative models, semantic segmentation and image captioning with vision and language models. Based on this framework, we introduce ARAS400k: A large-scale Remote sensing dataset Augmented with Synthetic data for segmentation and captioning, containing 100k real images and 300k synthetic images, each paired with segmentation maps and descriptions.

ARAS400k enables the automated evaluation of synthetic data by analyzing semantic composition, minimizing caption redundancy, and verifying cross-modal consistency between visual structures and language descriptions. Experimental results indicate that while models trained exclusively on synthetic data reach competitive performance levels, those trained with augmented data (a combination of real and synthetic images) consistently outperform real-data baselines. Consequently, this work establishes a scalable benchmark for remote sensing tasks, specifically in semantic segmentation and image captioning. The dataset is available at \href{https://zenodo.org/records/18890661}{zenodo.org/records/18890661} and the code base at \href{https://github.com/caglarmert/ARAS400k}{github.com/caglarmert/ARAS400k}.
\end{abstract}    
\section{Introduction}
\label{sec:intro}
Generative data augmentation is an effective alternative when collecting additional real data is costly or infeasible \cite{guo2024generative}. However, synthetic samples often lack explainability and interpretability \cite{chen2024comprehensive}. Moreover, existing evaluation metrics often fail to capture the semantic alignment between synthetic and real images \cite{bandi2023power} and their measurement procedures are not transparent \cite{betzalel2024evaluation}.

\begin{figure}[ht]
    \centering
    \includegraphics[trim={20 120 370 0},clip, width=0.85\linewidth]{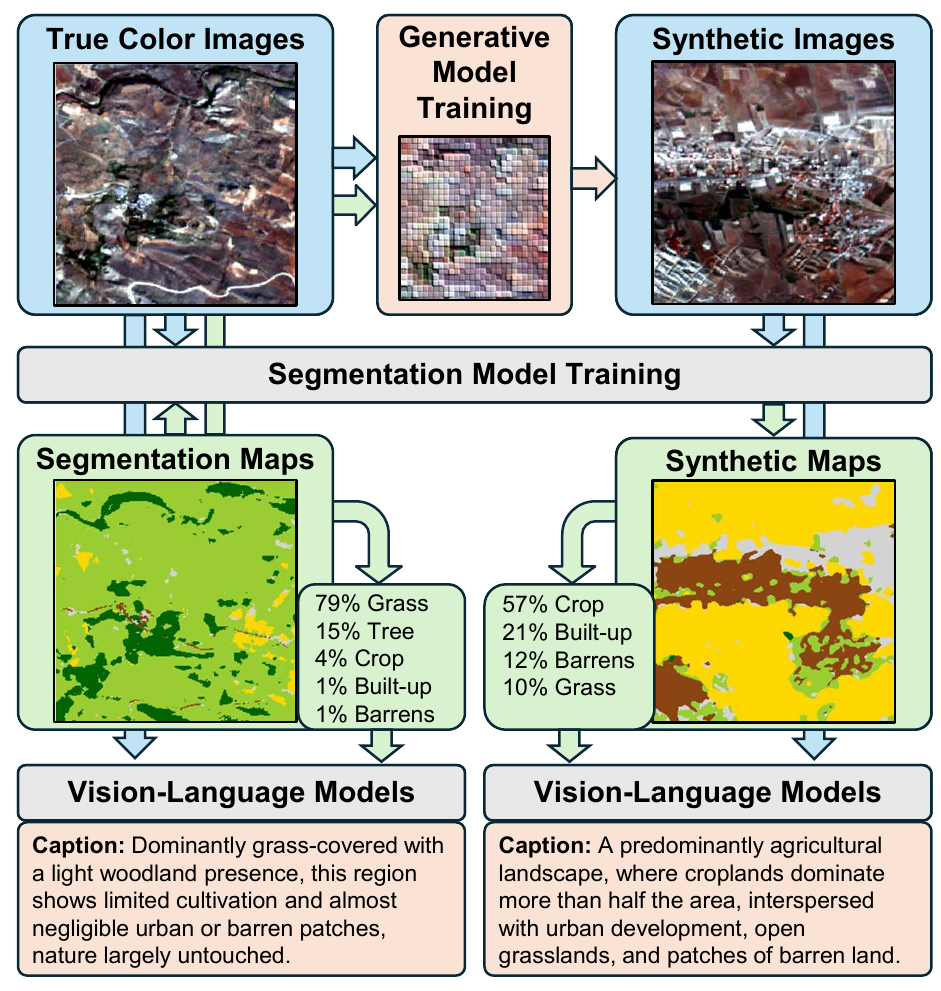}
    \caption{ARAS400k synthetic data augmentation framework.}
    \label{fig:generative_data_augmentation}
\end{figure}

To address these limitations, we propose a three-stage workflow (\Cref{fig:generative_data_augmentation}). In the first stage, we acquire and align true-color images with land cover data. After cleaning and pre-processing we produce an initial set of real image-segmentation map pairs. These pairs are then used to train semantic segmentation models, while a generative model is trained on the true-color images to generate synthetic samples. In the second stage, we utilize the segmentation models and create the segmentation maps of the synthetic samples. In the final stage we employ foundation models to generate descriptive captions by integrating visual content with composition statistics derived from the segmentation maps. 

\noindent The contributions of this work are summarized as follows:
\begin{itemize}
    \item \textbf{Large-Scale Multi-Modal Dataset:} We introduce the fully open-sourced and publicly available ARAS400k, a comprehensive remote sensing dataset consisting of 100,240 real images and 300,000 synthetic images, each paired with semantic segmentation maps and over 2 million descriptive captions. 
    \item \textbf{Automated Context-Aware Framework:} We propose an automated pipeline for context-aware caption generation and evaluation that utilizes composition statistics available from ground-truth or obtainable from segmentation models, reducing reliance on manual labor.
    \item \textbf{Vision-Language Integration:} We demonstrate a novel integration of foundation models to guide synthetic data evaluation through semantic consistency and redundancy reduction.
    \item \textbf{Proven Downstream Utility:} We provide an experimental benchmark showing that models trained on our augmented dataset (real + synthetic) consistently outperform those trained on real data alone, particularly in addressing class-imbalance for under-represented categories.
\end{itemize}

\section{Related Works}
\label{sec:related_works}
Recent advances demonstrate that synthetic data can effectively substitute or augment real datasets. SynthCLIP~\cite{hammoud2024synthclip} and SynGround~\cite{he2024learning} show that models trained exclusively on synthetic image-caption pairs can achieve performance comparable to real-data models in visual grounding and alignment tasks. The utility of such data is further enhanced by distillation techniques that prioritize high-quality samples over volume~\cite{wang2024not}. 

In the remote sensing domain, the need for precise spatial alignment and multi-scale feature integration is a challenge. Combining detail attention sampling with a teacher-student network effectively integrates local and global features, yielding state-of-the-art accuracy in high-resolution remote sensing image classification~\cite{liu2025remote}. Research into multidimensional datasets suggests that the strategic sequencing of Generative Adversarial Networks (GANs) and feature selection is critical for maintaining model validity and generalization capability~\cite{al2023synthetic}. 

Although Denoising Diffusion Probabilistic Models~\cite{ho2020denoising} and newer variants have proven to be successful in generating images, they often require longer training and inference times compared to GAN architectures~\cite{yazdani2025generative}. On the other hand, GAN models tend to have problems such as mode collapse, vanishing gradients, non converging or unstable training and they are sensitive to hyperparameters~\cite{barsha2025depth}. Various methodologies have been developed to address these limitations by improving convergence and output quality~\cite{huang2024gan}. Recent advancements also explore hybrid approaches that combine the fast inference of GANs with the high fidelity of diffusion models through distillation techniques~\cite{kang2024distilling}.

Finally, evaluating the large scale datasets and generative models requires more than standard metrics. The CLIPScore~\cite{hessel2021clipscore} metric is shown to align more with human assessment, enabling reference-free caption evaluation. Although it is especially crucial for non-referenced synthetic data, the ultimate validation for synthetic data lies in the ability to improve downstream tasks, such as semantic segmentation.
\section{Methods}
\label{sec:methods}
\subsection{Data Acquisition}
\label{subsec:data_acquisition}
We acquired data from ESA Sentinel-2 RGBNIR true-color images and WorldCover 2021~\cite{zanaga2022esa} land cover maps. We then geo-aligned the land cover and true-color layers using their respective geographic coordinates and extracted 157,283 image patches at a resolution of $256 \times 256$ pixels.

The original land cover data contains 11 classes: tree, shrub, grass, crop, built-up, barren, snow, water, wetland, mangrove and moss. However, several classes are underrepresented. Snow, wetland, moss and mangrove together appear only in 1,719 patches (1.1\%). To address this imbalance, we merged snow into barren, mangrove and wetland into tree cover and moss into grass, resulting in seven final classes. 

Patches dominated by water often had sun glare, which negatively affect image color and brightness, thus we removed 7,063 patches containing more than 90\% water coverage. After filtering out these samples and removing 49,980 patches with missing values, we ended up with a total of 100,240 patches covering an area of $656{,}933\,\mathrm{km}^2$. 

\subsection{Data Generation Pipeline}

\begin{figure}[ht]
    \centering
    \includegraphics[trim={50 0 50 0}, clip, width=0.9\linewidth]{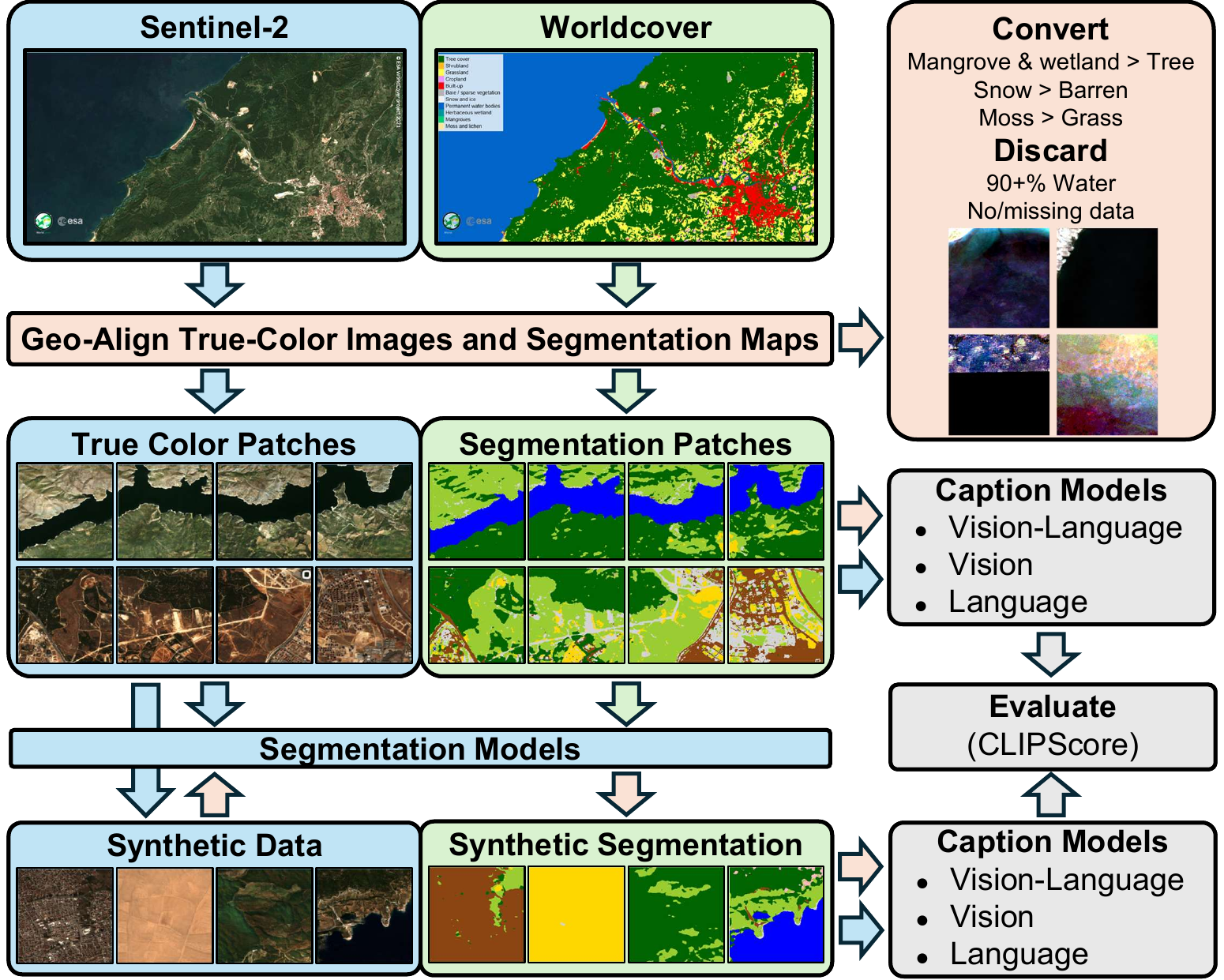}
    \caption{Data generation pipeline, comprised of data acquisition, alignment, cleaning, pre-processing, segmentation and generative model training, caption generation and evaluation.}
    \label{fig:data_gen_pipeline}
\end{figure}

We develop a cross-domain pipeline, illustrated in \Cref{fig:data_gen_pipeline}, integrating visual and textual information to improve interpretability and contextual grounding. The pipeline performs generative data augmentation to increase data diversity and volume, and uses image captioning to translate visual content into language descriptions. This enables language models to evaluate both real and synthetic images using semantic and contextual information.

We updated the official implementation of the Alias-Free Generative Adversarial Networks (StyleGAN3)~\cite{Karras2021} to be more efficient, utilizing the updated libraries and newer GPU architectures, enabling faster training and inference. With this implementation, we used real true-color images to train our StyleGAN3 model from scratch. We also combined the ideas from Semantic Image Synthesis with Spatially-Adaptive Normalization (SPADE)~\cite{park2019SPADE} and U-Net Based Discriminator for Generative Adversarial Networks~\cite{schonfeld2020u} to train our custom segmentation map conditioned GAN architecture. We used real images and corresponding segmentation maps to generate synthetic augmentations of existing image-segmentation pairs.

Our framework uses fine-tuned, trained-from-scratch or pre-trained models for the following tasks:
\begin{itemize}
    \item \textbf{Image captioning:} Gemma3~\cite{team2025gemma}, Qwen3-VL~\cite{qwen3technicalreport} generate descriptive captions from images, enabling semantic interpretation through language representations.
    \item \textbf{Image generation:} Our customized models based on StyleGAN3~\cite{Karras2021}, SPADE~\cite{park2019SPADE} and U-Net based discriminator~\cite{schonfeld2020u} produces synthetic images to augment the dataset.
    \item \textbf{Image segmentation:} UNet~\cite{ronneberger2015u}, UNet++~\cite{zhou2018unet++}, PAN~\cite{li2018pyramid}, DeepLabV3+~\cite{chen2018encoder}, SegFormer~\cite{xie2021segformer} and FPN~\cite{lin2017feature} generate semantic segmentation maps for scene understanding. 
    \item \textbf{Composition description:} Gemma3~\cite{team2025gemma} and Qwen3~\cite{qwen3technicalreport} convert scene composition into textual descriptions.
\end{itemize}
\section{ARAS400k}
\label{sec:a_new_dataset}
The proposed framework for ARAS400k (A large-scale Remote sensing dataset Augmented with Synthetic data for segmentation and captioning), combines the strengths of vision and language domains. Specifically, semantic segmentation to process remote sensing images, providing structural representation of images with predetermined land cover classes. Composition of a scene from the outputs of the semantic segmentation is obtained to present a high-level summary. Remote sensing images, segmentation maps and compositions are used by language models to generate captions.

To demonstrate the effectiveness of our framework, we collected, cleaned and pre-processed remote sensing image and land cover data, as detailed in \Cref{subsec:data_acquisition} and shown in \Cref{fig:data_gen_pipeline}. Then we trained segmentation models from real image-segmentation pairs and generative models from real images. We utilized our generative models to generate synthetic samples, as shown in \Cref{fig:rssg_grid}, effectively quadrupling the volume of the existing dataset and segmentation models to obtain segmentation maps. Finally, we used pre-trained open-sourced foundation models for descriptive caption generation, using only images, composition statistics or both. 

We used CLIPScore~\cite{hessel2021clipscore}, the reference-free caption evaluation metric, to assess the alignment of the generated captions. We also calculated caption redundancy of the existing and our datasets, by finding unique strings in each dataset.

To improve downstream tasks, we introduced synthetic samples for segmentation model training to demonstrate the usefulness of the synthetic samples. This synthetic data augmentation effectively improved segmentation performance for all class-based and overall performance metrics. To prevent data-leakage, and enable reproducibility, we fixed train, validation and test sets and used only train set to generate synthetic samples. 

\begin{table}[ht]
    \centering
    \small
    \setlength{\tabcolsep}{3pt}
\renewcommand{\arraystretch}{1.1}
    \begin{tabular}{lrrrr}
        \toprule
        \textbf{Dataset} & \textbf{Volume} & \textbf{Captions}  & \textbf{Redundancy} & \textbf{CLIPScore}\\
        \midrule
        NWPU         & 31,500    & 157,500    & 72.65\% & 30.25 \\
        RSICD        & 10,921    & 54,605     & 67.02\% & 29.11 \\
        UCMC         & 2,100     & 10,500     & 80.88\% & 30.18 \\
        ARAS400k     & 400,240   & 2,001,200& 12.85\% & 29.66 \\
        ~~~~~~-Real  & 100,240   & 501,200  & 8.01\%  & 29.89 \\
        ~~~~~~-Synth & 300,000   & 1,500,000& 13.06\% & 29.58 \\
        \bottomrule       
    \end{tabular}
    \caption{Comparison of remote sensing image caption datasets in terms of total number of images (volume), captions and redundancy.}
    \label{tab:dataset_comparison}    
\end{table}

As shown in \Cref{tab:dataset_comparison}, the volume of our dataset, with 100,240 real and 300,000 synthetic images and a total of 2,001,200 captions, significantly exceeds existing remote sensing benchmarks such as NWPU~\cite{cheng2022nwpu} containing 31,500, RSICD~\cite{lu2017exploring} with 10,921, and UCMC~\cite{qu2016deep} with 2,100. Furthermore, our dataset is much richer in caption variety. While traditional datasets like UCMC and NWPU show high redundancy rates of 80.88\% and 72.65\% respectively, ARAS400k achieves a significantly lower redundancy of 12.85\%. This demonstrates that while previously introduced remote sensing caption datasets facilitate image-to-text training, they are relatively limited in scale and often contain repetitive captions introduced by human annotators.

\begin{table}[ht]
\centering
\small
\setlength{\tabcolsep}{2pt}
\renewcommand{\arraystretch}{1.2}
\begin{tabular}{lcccc}
\toprule
\textbf{Model} & \textbf{Method} & \textbf{Unique} & \textbf{Redundancy} & \textbf{CLIPScore}\\
\textbf{ARAS400k} &&& \\
Qwen3-4B                & Text   & 214,319 & 46,45   \% & 26.34 \\
Gemma3-4B               & Vision & 396,878 & 0.84    \% & 31.06 \\
Qwen3-VL-8B             & Vision & 359,997 & 10.05   \% & 31.79 \\
Gemma3-4B               & Hybrid & 398,339 & 0.47    \% & 30.26 \\
Qwen3-VL-8B             & Hybrid & 374,570 & 6.41    \% & 27.88 \\
\midrule
\textbf{Real 100k} &&& \\
Qwen3-4B                & Text   & 71,899  & 28,27   \% & 26.48 \\
Gemma3-4B               & Vision & 99,981  & 0.26    \% & 30.60 \\
Qwen3-VL-8B             & Vision & 91,269  & 8.95    \% & 30.67 \\
Gemma3-4B               & Hybrid & 100,105 & 0.13    \% & 29.90 \\
Qwen3-VL-8B             & Hybrid & 97,814  & 2.42    \% & 27.87 \\
\midrule
\textbf{Synthetic 300k} &&& \\
Qwen3-4B                & Text   & 159,138 & 46.95   \% & 26.29 \\
Gemma3-4B               & Vision & 296,900 & 1.03    \% & 31.21 \\
Qwen3-VL-8B             & Vision & 268,821 & 10.39   \% & 32.16 \\
Gemma3-4B               & Hybrid & 298,556 & 2.70    \% & 30.38 \\
Qwen3-VL-8B             & Hybrid & 280,731 & 6.42    \% & 27.88 \\
\bottomrule
\end{tabular}
\caption{Captions generated by each model, input modality, unique sentences and the redundancy rates in our dataset.}
\label{tab:model-comparison-transposed}
\end{table}

Our captioning approach employ three distinct modalities using foundation models.

\begin{itemize}
    \item Text: Converts segmentation-derived composition statistics into descriptions.
    \item Vision: Generates captions directly from image content.
    \item Hybrid: Integrates both visual content and composition statistics. Employs vision-language foundation models to generate captions and language models to refine percentage based composition statistics obtained from the segmentation maps.
\end{itemize}

Detailed comparisons in \Cref{tab:model-comparison-transposed} show that hybrid methods achieve better performance and the lowest redundancy across both real and synthetic data subsets. While real-data subsets exhibit lower baseline redundancy, all configurations follow a consistent pattern: hybrid models yield the highest variety, whereas text-only result in the highest redundancy.

\begin{figure}[ht]
    \centering
    \includegraphics[trim={10 10 10 10}, clip, width=1\linewidth]{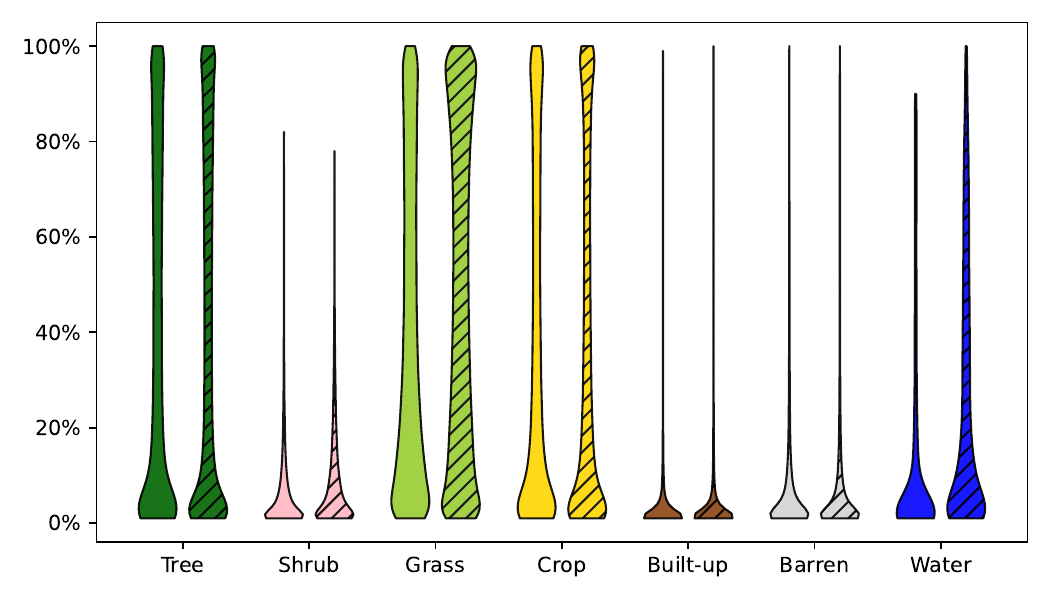}
    \caption{Real and synthetic (hatched) per-class composition statistics, indicating strong similarity between real and synthetic data.}
    \label{fig:rss2021_real_synth_violin}
\end{figure}

We also provide image segmentation models trained with the image-segmentation pairs. Finally, we trained generative models for synthetic data augmentation to increase the volume and variety of the existing real data. We used our segmentation models to obtain segmentation maps of our synthetic data and obtained composition statistics, similar to real data as seen from the \Cref{fig:rss2021_real_synth_violin}. 


\begin{figure}[ht]
    \centering
    \includegraphics[width=1\linewidth]{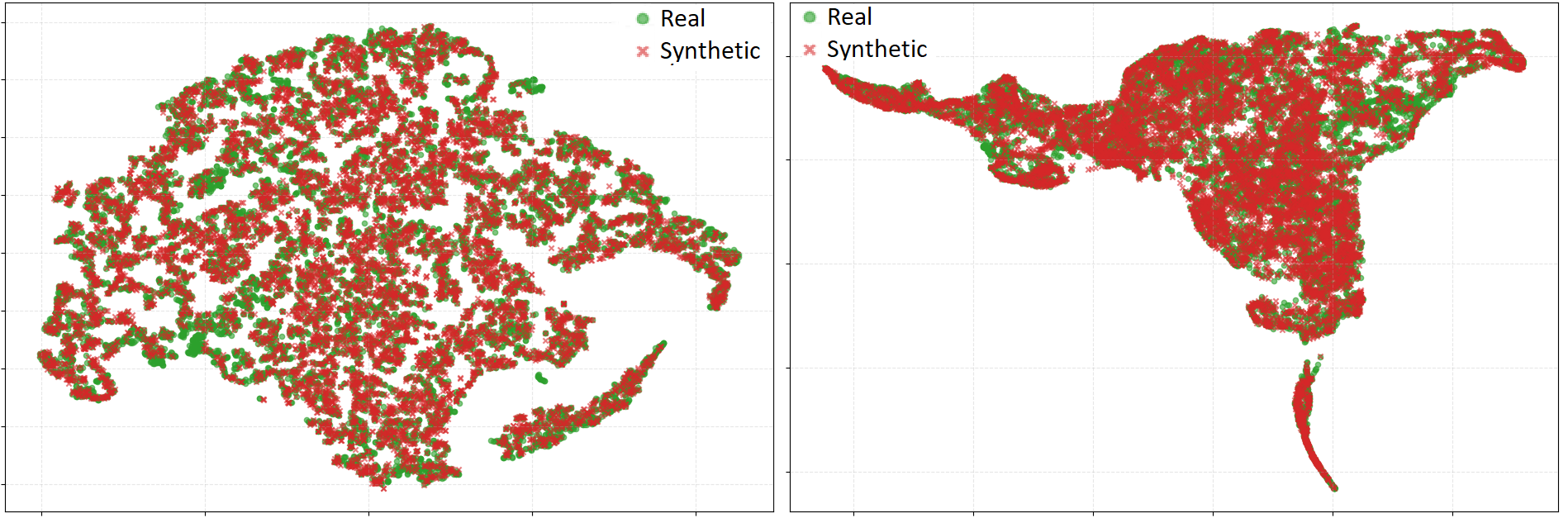}
    \caption{t-SNE (left) and UMAP (right) comparison of real and synthetic samples, showing similarity of real and synthetic data.}
    \label{fig:t-SNE}
\end{figure}

We also compared t-distributed Stochastic Neighbor Embedding (t-SNE) and Uniform Manifold Approximation and Projection (UMAP) of real and synthetic samples, using our trained segmentation model's feature extraction capability. The tightly grouped projection of real and synthetic samples inside similar clusters, as shown in the \Cref{fig:t-SNE}, indicate our synthetic samples are visually similar to the real data.

\begin{table*}[ht]
\centering
\small
\setlength{\tabcolsep}{3pt}
\begin{tabular}{lccccc|ccccc|ccccc}
\toprule
& \multicolumn{5}{c}{\textbf{Real Data}}
& \multicolumn{5}{c}{\textbf{Uncond. Synthetic 100k}}
& \multicolumn{5}{c}{\textbf{Uncond. Synthetic 300k}} \\
\cmidrule(lr){2-6}\cmidrule(lr){7-11}\cmidrule(lr){12-16}
\textbf{Model}
& F1 & IoU & Prec. & Recall & Acc.
& F1 & IoU & Prec. & Recall & Acc.
& F1 & IoU & Prec. & Recall & Acc. \\
\midrule
Unet      & 76.37 & 64.49 & 77.05 & 75.86 & 85.17 & 73.06 & 60.79 & \textbf{78.17} & 69.71 & 83.24 & 74.63 & 62.31 & 77.60 & 72.23 & 83.85 \\
Unet++    & 76.38 & 64.36 & 76.50 & 76.39 & 84.85 & \textbf{73.57} & \textbf{61.27} & 77.84 & \textbf{70.57} & 83.35 & 74.94 & 62.73 & 77.81 & 72.56 & 84.28 \\
PAN       & 76.20 & 64.10 & 75.29 & \textbf{77.60} & 84.62 & 73.07 & 60.71 & 77.53 & 70.04 & 83.19 & 74.65 & 62.42 & 77.70 & 72.22 & 84.14 \\
DeepLab   & 76.02 & 63.91 & 75.59 & 76.77 & 84.37 & 72.43 & 59.91 & 77.08 & 69.23 & 82.51 & 74.63 & 62.39 & 77.69 & 72.15 & 84.09 \\
Segformer & \textbf{77.09} & \textbf{65.26} & \textbf{78.10} & 76.33 & \textbf{85.66} & 73.48 & 61.23 & 77.77 & 70.40 & \textbf{83.63} & \textbf{75.10} & \textbf{62.90} & 77.65 & \textbf{72.95} & \textbf{84.33} \\
FPN       & 76.85 & 64.92 & 77.07 & 76.76 & 85.17 & 73.36 & 61.01 & 77.30 & 70.43 & 83.36 & 74.81 & 62.59 & \textbf{77.99} & 72.26 & 84.26 \\
\midrule
\textbf{Mean}
& 76.49 & 64.51 & 76.60 & \textbf{76.62} & 84.97
& 73.16 & 60.82 & 77.62 & 70.06 & 83.21
& 74.79 & 62.56 & 77.74 & 72.40 & 84.16 \\
\midrule\midrule

& \multicolumn{5}{c}{\textbf{Real + Uncond. 300k + Cond. 80k}}
& \multicolumn{5}{c}{\textbf{Real + Cond. 80k}}
& \multicolumn{5}{c}{\textbf{Real + Uncond. 300k}} \\
\cmidrule(lr){2-6}\cmidrule(lr){7-11}\cmidrule(lr){12-16}
\textbf{Model}
& F1 & IoU & Prec. & Recall & Acc.
& F1 & IoU & Prec. & Recall & Acc.
& F1 & IoU & Prec. & Recall & Acc. \\
\midrule
Unet      & 76.36 & 64.52 & 79.41 & 74.00 & 85.43 & 78.00 & 66.39 & 79.11 & \textbf{77.04} & 86.20 & 77.40 & 65.58 & 79.37 & 75.64 & 85.79 \\
Unet++    & \textbf{77.70} & \textbf{66.04} & \textbf{79.70} & \textbf{75.96} & 86.08 & \textbf{77.98} & \textbf{66.40} & \textbf{79.52} & 76.70 & 86.19 & \textbf{78.23} & \textbf{66.67} & 80.05 & \textbf{76.65} & \textbf{86.44} \\
PAN       & 77.29 & 65.52 & 79.21 & 75.64 & 85.86 & 76.00 & 63.90 & 77.00 & 75.17 & 84.50 & 77.21 & 65.48 & 79.94 & 75.02 & 86.04 \\
DeepLab   & 77.18 & 65.42 & 79.48 & 75.27 & 85.91 & 77.59 & 65.87 & 78.33 & 76.91 & 85.83 & 77.62 & 65.94 & 80.00 & 75.65 & 86.20 \\
Segformer & 77.43 & 65.72 & 79.58 & 75.61 & \textbf{86.10} & 77.91 & 66.27 & 79.09 & 76.83 & \textbf{86.20} & 77.80 & 66.14 & 79.98 & 75.98 & 86.30 \\
FPN       & 77.45 & 65.74 & 79.51 & 75.68 & 86.08 & 77.88 & 66.30 & 79.57 & 76.43 & 86.29 & 77.71 & 66.04 & \textbf{80.16} & 75.66 & 86.28 \\
\midrule
\textbf{Mean}
& 77.24 & 65.49 & 79.48 & 75.36 & 85.91
& 77.56 & 65.86 & 78.77 & 76.51 & 85.87
& \textbf{77.66} & \textbf{65.98} & \textbf{79.92} & 75.77 & \textbf{86.18} \\
\bottomrule
\end{tabular}
\caption{Overall test performance comparison of segmentation architectures trained on real and synthetic data (values in \%).}
\label{tab:stacked_overall_results}
\end{table*}

\begin{table*}[ht]
\small
\centering
\setlength{\tabcolsep}{3pt}
\begin{tabular}{lccccccc|ccccccc}
\toprule
& \multicolumn{7}{c|}{\textbf{Real Data}}
& \multicolumn{7}{c}{\textbf{Uncond. Synthetic 100k}} \\
\cmidrule(lr){2-8}\cmidrule(lr){9-15}
\textbf{Model}
& Tree & Shrub & Grass & Crop & Build & Barren & Water
& Tree & Shrub & Grass & Crop & Build & Barren & Water \\
\midrule
Unet
& 90.92 & 45.86 & 84.25 & 86.82 & 73.86 & 58.98 & 93.88
& 89.80 & 35.65 & 82.70 & 83.93 & 71.64 & 55.84 & 91.88 \\
Unet++
& 90.88 & 45.50 & 83.95 & 86.30 & \textbf{74.11} & 60.97 & 92.93
& \textbf{90.04} & \textbf{37.64} & 82.82 & 83.90 & \textbf{71.85} & \textbf{56.66} & 92.05 \\
PAN
& 90.31 & 46.88 & 84.05 & 86.38 & 71.50 & 60.58 & 93.69
& 89.73 & 36.07 & 82.74 & 83.90 & 70.94 & 56.57 & 91.54 \\
DeepLab
& 90.42 & 46.22 & 83.56 & 86.24 & 72.21 & 59.92 & 93.58
& 89.61 & 36.40 & 81.97 & 82.66 & 70.13 & 54.76 & 91.50 \\
Segformer
& \textbf{91.08} & \textbf{48.80} & \textbf{85.07} & \textbf{87.16} & 73.05 & 60.25 & \textbf{94.20}
& 90.04 & 37.40 & \textbf{83.08} & \textbf{84.51} & 71.13 & 56.14 & \textbf{92.06} \\
FPN
& 90.40 & 47.67 & 84.48 & 87.12 & 73.02 & \textbf{61.21} & 94.05
& 89.87 & 38.01 & 82.85 & 84.11 & 71.01 & 55.74 & 91.90 \\
\midrule
\textbf{Mean}
& 90.67 & 46.82 & 84.23 & 86.67 & 72.96 & 60.32 & 93.72
& 89.85 & 36.86 & 82.69 & 83.83 & 71.12 & 55.95 & 91.82 \\
\bottomrule
\end{tabular}
\begin{tabular}{lccccccc|ccccccc}
\toprule
& \multicolumn{7}{c|}{\textbf{Uncond. Synthetic 300k}}
& \multicolumn{7}{c}{\textbf{Real + Uncond. 300k + Cond. 80k}} \\
\cmidrule(lr){2-8}\cmidrule(lr){9-15}
\textbf{Model}
& Tree & Shrub & Grass & Crop & Build & Barren & Water
& Tree & Shrub & Grass & Crop & Build & Barren & Water \\
\midrule
Unet
& 90.31 & 42.23 & 83.33 & 84.53 & \textbf{72.16} & 57.61 & 92.22
& 91.02 & 45.44 & 84.64 & 86.75 & 74.03 & 58.87 & 93.75 \\
Unet++
& 90.40 & 42.73 & 83.73 & \textbf{85.38} & 72.05 & 57.85 & 92.44
& 91.47 & \textbf{48.17} & 85.21 & 87.56 & \textbf{74.74} & \textbf{62.54} & \textbf{94.19} \\
PAN
& 90.34 & 41.73 & 83.61 & 85.18 & 71.83 & 57.63 & 92.25
& 91.29 & 48.06 & 85.04 & 87.36 & 73.50 & 61.74 & 94.03 \\
DeepLab
& 90.35 & 41.71 & 83.56 & 85.05 & 71.90 & 57.58 & 92.27
& 91.39 & 47.82 & 85.13 & 87.30 & 73.69 & 60.92 & 94.01 \\
Segformer
& 90.43 & \textbf{43.35} & \textbf{83.85} & 85.37 & 72.06 & \textbf{58.01} & \textbf{92.63}
& 91.47 & 48.07 & \textbf{85.29} & \textbf{87.58} & 73.59 & 61.87 & 94.12 \\
FPN
& \textbf{90.44} & 42.39 & 83.76 & 85.19 & 71.81 & 57.71 & 92.39
& \textbf{91.50} & \textbf{48.17} & 85.19 & \textbf{87.58} & 73.64 & 61.94 & 94.11 \\
\midrule
\textbf{Mean}
& 90.38 & 42.36 & 83.64 & 85.12 & 71.97 & 57.73 & 92.37
& 91.36 & 47.96 & 85.08 & 87.36 & 73.87 & 61.31 & 94.04 \\
\bottomrule
\end{tabular}
\begin{tabular}{lccccccc|ccccccc}
\toprule
& \multicolumn{7}{c|}{\textbf{Real + Cond. 80k}}
& \multicolumn{7}{c}{\textbf{Real + Uncond. 300k}} \\
\cmidrule(lr){2-8}\cmidrule(lr){9-15}
\textbf{Model}
& Tree & Shrub & Grass & Crop & Build & Barren & Water
& Tree & Shrub & Grass & Crop & Build & Barren & Water \\
\midrule
Unet
& 91.58 & 48.94 & 85.41 & 87.58 & 74.92 & 63.14 & 94.42
& 91.42 & 48.59 & 84.89 & 87.23 & 73.66 & 62.29 & 93.69 \\
Unet++
& 91.59 & 48.35 & 85.46 & 87.54 & \textbf{75.36} & \textbf{63.27} & 94.32
& \textbf{91.77} & \textbf{49.75} & \textbf{85.61} & \textbf{87.83} & \textbf{75.18} & \textbf{63.18} & \textbf{94.31} \\
PAN
& 90.25 & 45.69 & 83.26 & 86.35 & 72.31 & 60.35 & 93.78
& 91.38 & 47.61 & 85.24 & 87.53 & 73.20 & 61.53 & 94.00 \\
DeepLab
& 91.36 & 48.81 & 85.05 & 87.24 & 73.95 & 62.26 & 94.43
& 91.53 & 48.88 & 85.37 & 87.69 & 73.79 & 61.93 & 94.15 \\
Segformer
& 91.58 & \textbf{49.20} & 85.47 & \textbf{87.65} & 74.11 & 62.97 & 94.36
& 91.57 & 48.99 & 85.48 & 87.80 & 73.75 & 62.85 & 94.16 \\
FPN
& \textbf{91.72} & 48.67 & \textbf{85.61} & 87.54 & 74.31 & 62.62 & \textbf{94.65}
& 91.55 & 48.94 & 85.51 & 87.74 & 73.49 & 62.54 & 94.19 \\
\midrule
\textbf{Mean}
& 91.35 & \textbf{48.94} & 85.04 & 87.32 & \textbf{74.16} & \textbf{62.44} & \textbf{94.33}
& \textbf{91.54} & 48.79 & \textbf{85.35} & \textbf{87.64} & 73.85 & 62.05 & 94.08 \\
\bottomrule
\end{tabular}
\caption{Per-class F1-score comparison of segmentation models (values in \%).}
\label{tab:stacked_perclass_real_synth}
\end{table*}

\section{Experimental Evaluation}
\label{sec:results}
\begin{figure}[ht]
\centering
\small
\setlength{\tabcolsep}{1pt}
\begin{tabular}{cccc}
\textbf{Train} & \textbf{Mask} & \textbf{Conditioned} & \textbf{Unconditioned} \\
\includegraphics[width=0.115\textwidth]{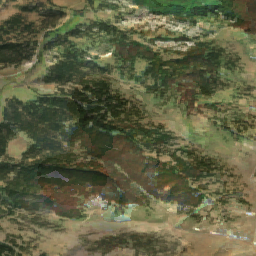} &
\includegraphics[width=0.115\textwidth]{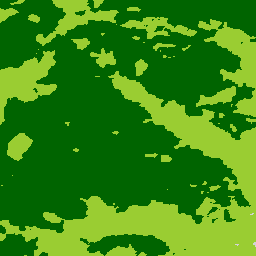} &
\includegraphics[width=0.115\textwidth]{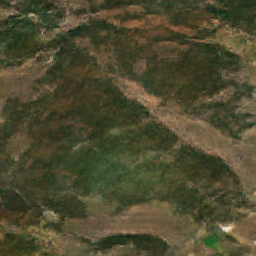} &
\includegraphics[width=0.115\textwidth]{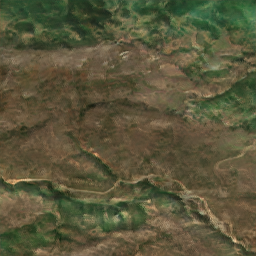} \\
\includegraphics[width=0.115\textwidth]{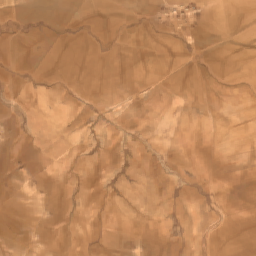} &
\includegraphics[width=0.115\textwidth]{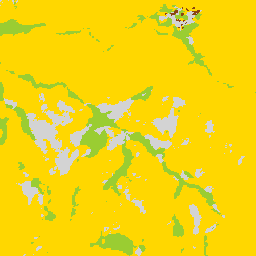} &
\includegraphics[width=0.115\textwidth]{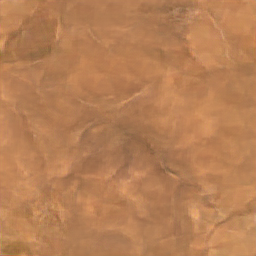} &
\includegraphics[width=0.115\textwidth]{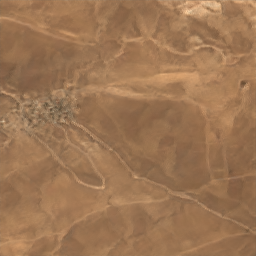} \\
\includegraphics[width=0.115\textwidth]{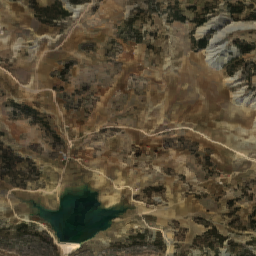} &
\includegraphics[width=0.115\textwidth]{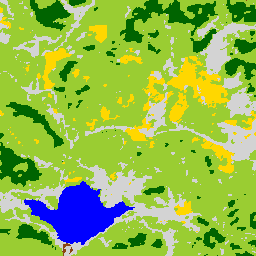} &
\includegraphics[width=0.115\textwidth]{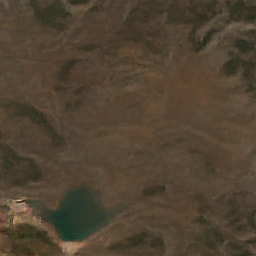} &
\includegraphics[width=0.115\textwidth]{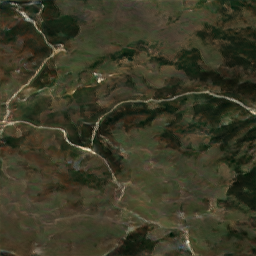} \\
\includegraphics[width=0.115\textwidth]{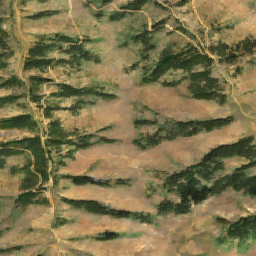} &
\includegraphics[width=0.115\textwidth]{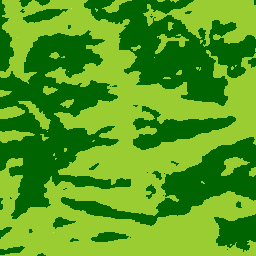} &
\includegraphics[width=0.115\textwidth]{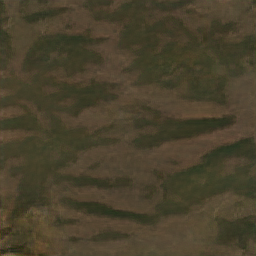} &
\includegraphics[width=0.115\textwidth]{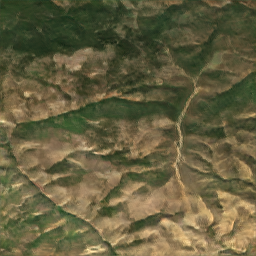} \\
\includegraphics[width=0.115\textwidth]{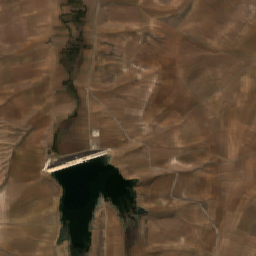} &
\includegraphics[width=0.115\textwidth]{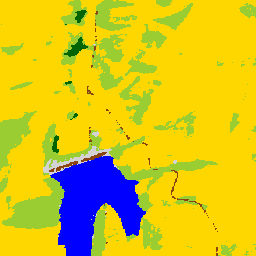} &
\includegraphics[width=0.115\textwidth]{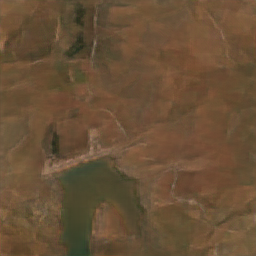} &
\includegraphics[width=0.115\textwidth]{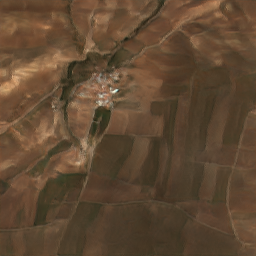} \\
\end{tabular}
\caption{Qualitative comparison of training images, masks, conditioned and unconditioned synthetic samples.}
\label{fig:rssg_grid}
\end{figure}

To ensure experimental rigor and prevent data leakage, the generative models were trained exclusively on a fixed training partition containing 80,182 real samples. Utilizing our implementation of the StyleGAN3~\cite{Karras2021}, U-Net based discriminator~\cite{schonfeld2020u} and SPADE~\cite{park2019SPADE} architectures, we trained our GAN models until their training FID score reached a plateau, indicating that the model had converged. Our updates to the StyleGAN3 training code-base, utilizing newer introductions to the libraries, enabled 44\% increase in the training and inference speed over the original official implementation. We present randomly selected real and synthetic samples in \Cref{fig:rssg_grid}, chosen for their visual similarity based on feature extraction from our pre-trained segmentation model. These comparisons show that the synthetic images are qualitatively similar to the real samples. FID scores for unconditional and conditional synthetic samples were measured as 16 and 72 respectively, indicating unconditional generation creates more realistic samples than the semantic segmentation mask conditioned generation. Furthermore, as shown in \Cref{tab:dataset_comparison}, existing datasets achieve CLIPScores between 29.11 and 30.25, meanwhile ARAS400k scores 29.89 and 29.58 CLIPScore for real and synthetic subsets respectively, indicating highly-competitive performance with human-annotated  benchmarks.

\Cref{tab:stacked_overall_results} presents the macro average F1 score, Intersection over Union (IoU), precision, recall and accuracy across all experimental configurations. The results of the models trained only with synthetic data are comparable yet slightly inferior to real data, indicating synthetic-only variants of our dataset is still capable to train solid models, with larger (300k) variant achieving better results than the smaller (100k). The augmented data configurations, consisting of both real and synthetic samples, achieve better results than the real data baseline, clearly indicating the advantage of synthetic data augmentation. Utilizing real data augmented with unconditional synthetic data samples (Real + Uncond. 300k) performs better than augmentation with conditional synthetic data (Real + Cond. 80k) or with both (Real + Uncond. 300k + Cond. 80k).

\Cref{tab:stacked_perclass_real_synth} presents a comprehensive per-class evaluation, reporting F1 scores for each land cover class. The results reveal substantial performance variation across categories: Water class achieves the highest scores, while shrubs pose a significant segmentation challenge, possibly due to scarcity of the shrub class and similarity with neighboring classes. Tree cover, croplands and grasslands were also successfully distinguished by the segmentation models, followed by built-up and barrens. Synthetic data provides comparable but slightly inferior results compared to the real data meanwhile the combination of real and synthetic data, shown as augmented data, achieves better performance across all classes consistently. The highest performance increase occurs in the least represented and lowest performing classes, indicating that synthetic data augmentation effectively alleviates class-imbalance problem.

\section{Discussion}
\label{sec:discussion}

We trained semantic segmentation models with real images, using different architectures, creating a baseline performance benchmark for this downstream task. To evaluate the contribution of conditional and unconditional synthetic data augmentation approaches, we used this baseline and re-trained same architectures with synthetic-augmented data. Details of this ablation study, shown in the \Cref{tab:stacked_overall_results}, highlight that using synthetic only data is an alternative to real data at the cost of a slight performance decrease. Using a comparable number of synthetic samples as the original real data, decreases the overall F1 score by 3.5 points. Meanwhile utilizing 3 times larger synthetic samples achieves a better result, trailing the real baseline by 2.0 F1 score. These results show that using synthetic samples instead of real data is feasible at a slight performance cost.

Injection of synthetic samples to real data improves the overall segmentation performance over the baseline. We tested three augmentation strategies: augmentation with segmentation map conditioned synthetic data, with unconditioned synthetic data, and a combination of both. Augmentation with unconditioned synthetic data slightly outperforms other strategies. The most important gain in using synthetic augmentation is more evident with under-represented and under-performing classes, as evident in the \Cref{tab:stacked_perclass_real_synth}. These results indicate that synthetic data augmentation improves downstream performance for the rare cases.

The results of these evaluations, summarized in the \Cref{tab:dataset_comparison}, indicate that our approach is much more diverse than existing datasets, while having between  $12-190\times$ more samples. Although CLIPScore for existing datasets are slightly higher than our dataset, some of our models performed better. As shown in the \Cref{tab:model-comparison-transposed}, vision models tend to provide more realistic descriptions in terms of CLIPScore. One distinct feature of our synthetic dataset over existing human-annotated datasets is the presence of land cover percentages. Our caption generation models, except for vision-only methods, had access to the composition statistics of the scene. We believe including this percentage information for generated captions is a valuable contribution, albeit the negative affects for the CLIPScore evaluation metric.

On a single NVIDIA H100 GPU, total processing and ablation study for this dataset spanned over 3,000 hours. A single end-to-end execution requires 80 hours for generative model training and synthetic sample generation, 20 hours for segmentation model training and inference, 300 hours for caption generation of 5 captions per each of 400,000 samples, totaling approximately 400 hours for whole dataset. Our framework processes 1,000 images per hour.

\section{Conclusion}
\label{sec:conclusion}
We introduced a vision-language framework for remote sensing synthetic data augmentation that integrates generative modeling, semantic segmentation, and automated image captioning. Using this framework, we produced ARAS400k, a large-scale dataset featuring over 100k real and 300k synthetic image-segmentation pairs. By utilizing foundation models to generate captions from both visual content and segmentation-derived composition statistics, our approach provides a scalable and interpretable solution for enhancing multi-modal remote sensing tasks. 

The experimental results for semantic segmentation establish a robust baseline with real data, while models trained exclusively on synthetic data achieved highly competitive performance. Most notably, models trained on augmented data (real and synthetic) achieved superior results, demonstrating the advantage of synthetic augmentation for scalable data expansion. Meanwhile reference-free CLIPScore evaluation confirm that our automated captions are comparable to human-annotated datasets while having significantly higher volume and lower redundancy. The consistent performance across real and synthetic subsets highlights the potential for using synthetic data as a viable alternative to sensitive or labor-intensive real-world datasets.

Possible future work includes improving synthetic image quality through remote sensing super-resolution approaches, developing specialized evaluation metrics for synthetic samples and exploring structure preserving image generation from segmentation maps. We have introduced this work as an example in remote sensing, due to the availability of large scale data acquisition through public data sources and we believe this work can be replicated for different domains, including but not limited to common scene understanding, autonomous driving and medical domains.

\section*{Acknowledgment}
This work has been supported by Middle East Technical University Scientific Research Projects Coordination Unit under grant number ADEP-704-2024-11486. The experiments reported in this work were fully performed at TUBITAK ULAKBIM, High Performance and Grid Computing Center (TRUBA).

{
    \small
    \bibliographystyle{ieeenat_fullname}
    \bibliography{main}

@String(CVPR= {IEEE Conf. Comput. Vis. Pattern Recog.})

@String(ECCV= {Eur. Conf. Comput. Vis.})

@String(NIPS= {Adv. Neural Inform. Process. Syst.})

@String(CVPR  = {CVPR})

@String(ECCV  = {ECCV})

@String(NIPS  = {NeurIPS})

@article{guo2024generative,
  title={Generative ai for synthetic data generation: Methods, challenges and the future},
  author={Guo, Xu and Chen, Yiqiang},
  journal={arXiv:2403.04190},
  year={2024}
}

@article{chen2024comprehensive,
  title={A comprehensive survey for generative data augmentation},
  author={Chen, Yunhao and Yan, Zihui and Zhu, Yunjie},
  journal={Neurocomputing},
  volume={600},
  pages={128167},
  year={2024},
  publisher={Elsevier}
}

@article{bandi2023power,
  title={The power of generative ai: A review of requirements, models, input--output formats, evaluation metrics, and challenges},
  author={Bandi, Ajay and Adapa, Pydi Venkata Satya Ramesh and Kuchi, Yudu Eswar Vinay Pratap Kumar},
  journal={Future Internet},
  volume={15},
  number={8},
  pages={260},
  year={2023},
  publisher={MDPI}
}

@article{betzalel2024evaluation,
  title={Evaluation metrics for generative models: An empirical study},
  author={Betzalel, Eyal and Penso, Coby and Fetaya, Ethan},
  journal={Mach. Learn. Knowl. Extr.},
  volume={6},
  number={3},
  year={2024},
}

@article{ho2020denoising,
  title={Denoising diffusion probabilistic models},
  author={Ho, Jonathan and Jain, Ajay and Abbeel, Pieter},
  journal={Advances in neural information processing systems},
  year={2020}
}

@article{lu2017exploring,
  title={Exploring models and data for remote sensing image caption generation},
  author={Lu, Xiaoqiang and Wang, Binqiang and Zheng, Xiangtao and Li, Xuelong},
  journal={IEEE Transactions on Geoscience and Remote Sensing},
  volume={56},
  number={4},
  pages={2183--2195},
  year={2017},
  publisher={IEEE}
}

@INPROCEEDINGS{qu2016deep,
  author={Qu, Bo and Li, Xuelong and Tao, Dacheng and Lu, Xiaoqiang},
  booktitle={2016 International Conference on Computer, Information and Telecommunication Systems (CITS)}, 
  title={Deep semantic understanding of high resolution remote sensing image}, 
  year={2016},
  volume={},
  number={},
  pages={1-5},
  doi={10.1109/CITS.2016.7546397}}

@article{cheng2022nwpu,
  title={NWPU-captions dataset and MLCA-net for remote sensing image captioning},
  author={Cheng, Qimin and Huang, Haiyan and Xu, Yuan and Zhou, Yuzhuo and Li, Huanying and Wang, Zhongyuan},
  journal={IEEE Transactions on Geoscience and Remote Sensing},
  volume={60},
  pages={1--19},
  year={2022},
  publisher={IEEE}
}

@article{team2025gemma,
  title={Gemma 3 technical report},
  author={Team, Gemma},
  journal={arXiv preprint arXiv:2503.19786},
  year={2025}
}

@misc{zanaga2022esa,
  title={ESA WorldCover 10 m 2021 v200},
  author={Zanaga, Daniele and Van De Kerchove, Ruben and Daems, Dirk and De Keersmaecker, Wanda and Brockmann, Carsten and Kirches, Grit and Wevers, Jan and Cartus, Oliver and Santoro, Maurizio and Fritz, Steffen and others},
  year={2022},
  url = {https://doi.org/10.5281/zenodo.7254221},
  doi = {10.5281/zenodo.7254220}
}

@misc{hammoud2024synthclip,
      title={SynthCLIP: Are We Ready for a Fully Synthetic CLIP Training?}, 
      author={Hasan Abed Al Kader Hammoud and Hani Itani and Fabio Pizzati and Philip Torr and Adel Bibi and Bernard Ghanem},
      year={2024},
      eprint={2402.01832},
      archivePrefix={arXiv},
}

@misc{he2024learning,
      title={Learning from Synthetic Data for Visual Grounding}, 
      author={Ruozhen He and Ziyan Yang and Paola Cascante-Bonilla and Alexander C. Berg and Vicente Ordonez},
      year={2024},
      eprint={2403.13804},
      archivePrefix={arXiv}
}

@article{wang2024not,
  title={Not all samples should be utilized equally: Towards understanding and improving dataset distillation},
  author={Wang, Shaobo and Yang, Yantai and Wang, Qilong and Li, Kaixin and Zhang, Linfeng and Yan, Junchi},
  journal={arXiv preprint arXiv:2408.12483},
  year={2024}
}

@inproceedings{ronneberger2015u,
  title={U-net: Convolutional networks for biomedical image segmentation},
  author={Ronneberger, Olaf and Fischer, Philipp and Brox, Thomas},
  booktitle={International Conference on Medical image computing and computer-assisted intervention},
  pages={234--241},
  year={2015},
  organization={Springer}
}

@inproceedings{zhou2018unet++,
  title={Unet++: A nested u-net architecture for medical image segmentation},
  author={Zhou, Zongwei and Rahman Siddiquee, Md Mahfuzur and Tajbakhsh, Nima and Liang, Jianming},
  booktitle={International workshop on deep learning in medical image analysis},
  year={2018},
}

@article{li2018pyramid,
  title={Pyramid attention network for semantic segmentation},
  author={Li, Hanchao and Xiong, Pengfei and An, Jie and Wang, Lingxue},
  journal={arXiv:1805.10180},
  year={2018}
}

@InProceedings{chen2018encoder,
author="Chen, Liang-Chieh
and Zhu, Yukun
and Papandreou, George
and Schroff, Florian
and Adam, Hartwig",
title="Encoder-Decoder with Atrous Separable Convolution for Semantic Image Segmentation",
booktitle="ECCV",
year="2018",
isbn="978-3-030-01234-2"
}

@article{xie2021segformer,
  title={SegFormer: Simple and efficient design for semantic segmentation with transformers},
  author={Xie, Enze and Wang, Wenhai and Yu, Zhiding and Anandkumar, Anima and Alvarez, Jose M and Luo, Ping},
  journal={Advances in neural information processing systems},
  volume={34},
  pages={12077--12090},
  year={2021}
}

@inproceedings{lin2017feature,
  title={Feature pyramid networks for object detection},
  author={Lin, Tsung-Yi and Doll{\'a}r, Piotr and Girshick, Ross and He, Kaiming and Hariharan, Bharath and Belongie, Serge},
  booktitle={Proceedings of the IEEE conference on computer vision and pattern recognition},
  pages={2117--2125},
  year={2017}
}

@inproceedings{hessel2021clipscore,
    title = "{CLIPS}core: A Reference-free Evaluation Metric for Image Captioning",
    author = "Hessel, Jack  and
      Holtzman, Ari  and
      Forbes, Maxwell  and
      Le Bras, Ronan  and
      Choi, Yejin",
    booktitle = "Conference on Empirical Methods in Natural Language Processing",
    year = "2021"
}

@article{yazdani2025generative,
  title={Generative AI in depth: A survey of recent advances, model variants, and real-world applications},
  author={Yazdani, Shamim and Singh, Akansha and Saxena, Nripsuta and Wang, Zichong and Palikhe, Avash and Pan, Deng and Pal, Umapada and Yang, Jie and Zhang, Wenbin},
  journal={Journal of Big Data},
  volume={12},
  number={1},
  pages={230},
  year={2025},
  publisher={Springer}
}

@article{liu2025remote,
  title={A remote sensing image classification method based on detail attention sampling and teacher-student network},
  author={Liu, Xinfu and Wu, Benze and Wu, Yirui},
  journal={ACM Journal of Data and Information Quality},
  volume={17},
  number={3},
  pages={1--19},
  year={2025},
  publisher={ACM New York, NY}
}

@article{al2023synthetic,
  title={Synthetic generation of multidimensional data to improve classification model validity},
  author={Al--Qerem, Ahmad and Ali, Ali Mohd and Attar, Hani and Nashwan, Shadi and Qi, Lianyong and Moghimi, Mohammad Kazem and Solyman, Ahmed},
  journal={ACM Journal of Data and Information Quality},
  volume={15},
  number={3},
  pages={1--20},
  year={2023},
  publisher={ACM New York, NY}
}

@article{barsha2025depth,
  title={An in-depth review and analysis of mode collapse in generative adversarial networks},
  author={Barsha, Farhat Lamia and Eberle, William},
  journal={Machine Learning},
  volume={114},
  number={6},
  pages={141},
  year={2025},
  publisher={Springer}
}

@article{huang2024gan,
  title={The {GAN} is dead; long live the {GAN}! a modern {GAN} baseline},
  author={Huang, Nick and Gokaslan, Aaron and Kuleshov, Volodymyr and Tompkin, James},
  journal={Advances in Neural Information Processing Systems},
  volume={37},
  pages={44177--44215},
  year={2024}
}

@InProceedings{kang2024distilling,
author="Kang, Minguk
and Zhang, Richard
and Barnes, Connelly
and Paris, Sylvain
and Kwak, Suha
and Park, Jaesik
and Shechtman, Eli
and Zhu, Jun-Yan
and Park, Taesung",
title="Distilling Diffusion Models Into Conditional {GANs}",
booktitle="ECCV",
year="2024",
isbn="978-3-031-73390-1"
}

@inproceedings{Karras2021,
author = {Karras, Tero and Aittala, Miika and Laine, Samuli and H\"{a}rk\"{o}nen, Erik and Hellsten, Janne and Lehtinen, Jaakko and Aila, Timo},
title = {Alias-free generative adversarial networks},
year = {2021},
booktitle = {International Conference on Neural Information Processing Systems},
series = {NIPS '21}
}

@INPROCEEDINGS{park2019SPADE,
  author={Park, Taesung and Liu, Ming-Yu and Wang, Ting-Chun and Zhu, Jun-Yan},
  booktitle={2019 IEEE/CVF Conference on Computer Vision and Pattern Recognition (CVPR)}, 
  title={Semantic Image Synthesis With Spatially-Adaptive Normalization}, 
  year={2019},
  volume={},
  number={},
  pages={2332-2341},
  doi={10.1109/CVPR.2019.00244}}

@inproceedings{schonfeld2020u,
  title={A {U-net} based discriminator for generative adversarial networks},
  author={Schonfeld, Edgar and Schiele, Bernt and Khoreva, Anna},
  booktitle={CVPR},
  year={2020}
}

@misc{qwen3technicalreport,
      title={Qwen3 Technical Report}, 
      author={Qwen Team},
      year={2025},
      eprint={2505.09388},
      archivePrefix={arXiv},
      primaryClass={cs.CL},
      url={https://arxiv.org/abs/2505.09388}, 
}
}


\end{document}